\documentclass[letterpaper]{article} 
\usepackage{aaai23}  
\usepackage{times}  
\usepackage{helvet}  
\usepackage{courier}  
\usepackage[hyphens]{url}  
\usepackage{graphicx} 
\usepackage{natbib}  
\usepackage{caption} 
\usepackage{algorithm}
\usepackage{algorithmic}
\usepackage{makecell}
\usepackage{tabu}
\usepackage{color}
\usepackage{newfloat}
\usepackage{listings}

\usepackage{tabularx}
\usepackage{microtype}
\usepackage{subfigure}
\usepackage{amsfonts,amssymb}
\usepackage{amsmath}
\usepackage{multirow}
\usepackage{courier}

\usepackage{newfloat}
\usepackage{listings}
\DeclareCaptionStyle{ruled}{labelfont=normalfont,labelsep=colon,strut=off} 
\floatstyle{ruled}
\newfloat{listing}{tb}{lst}{}
\floatname{listing}{Listing}
\title{Joint Multimodal Entity-Relation Extraction Based on Edge-enhanced Graph Alignment Network and Word-pair Relation Tagging}
\author{  
Li Yuan\textsuperscript{\rm 1,} \textsuperscript{\rm 2}, Yi Cai \textsuperscript{\rm1,} \textsuperscript{\rm2,} \textsuperscript{\rm3} \thanks{Corresponding author: Yi Cai (ycai@scut.edu.cn)}, Jin Wang\textsuperscript{\rm 4}, Qing Li\textsuperscript{\rm 5}
}
\affiliations{
    \textsuperscript{\rm 1}School of Software Engineering, South China University of Technology, Guangzhou, China\\
    \textsuperscript{\rm 2} Key Laboratory of Big Data and Intelligent Robot (SCUT), MOE of China\\
    \textsuperscript{\rm 3} The Peng Cheng Laboratory, Shenzhen, China. \\
    \textsuperscript{\rm 4} School of Information Science and Engineering, Yunnan University, Yunnan, P.R. China\\
    \textsuperscript{\rm 5} Department of Computing, The Hong Kong Polytechnic University, Hong Kong, China\\

  seyuanli@mail.scut.edu.cn, ycai@scut.edu.cn, wangjin@ynu.edu.cn, qing-prof.li@polyu.edu.hk
}

\usepackage{bibentry}

\begin{document}

\maketitle

\begin{abstract}
Multimodal named entity recognition (MNER) and multimodal relation extraction (MRE) are two fundamental subtasks in the multimodal knowledge graph construction task. However, the existing methods usually handle two tasks independently, which ignores the bidirectional interaction between them. This paper is the first to propose jointly performing MNER and MRE as a joint multimodal entity-relation extraction task (JMERE).
Besides, the current MNER and MRE models only consider aligning the visual objects with textual entities in visual and textual graphs but ignore the entity-entity relationships and object-object relationships. To address the above challenges, we propose an edge-enhanced graph alignment network and a word-pair relation tagging (EEGA) for the JMERE task. Specifically, we first design a word-pair relation tagging to exploit the bidirectional interaction between MNER and MRE and avoid error propagation. Then, we propose an edge-enhanced graph alignment network to enhance the JMERE task by aligning nodes and edges in the cross-graph. Compared with previous methods, the proposed method can leverage the edge information to auxiliary alignment between objects and entities and find the correlations between entity-entity relationships and object-object relationships. Experiments are conducted to show the effectiveness of our model\footnote{The code and appendix are available at https://github.com/YuanLi95/EEGA-for-JMERE}.

\end{abstract}
\section{Introduction}
\begin{figure}  
  \centering
    \includegraphics[scale=.68]{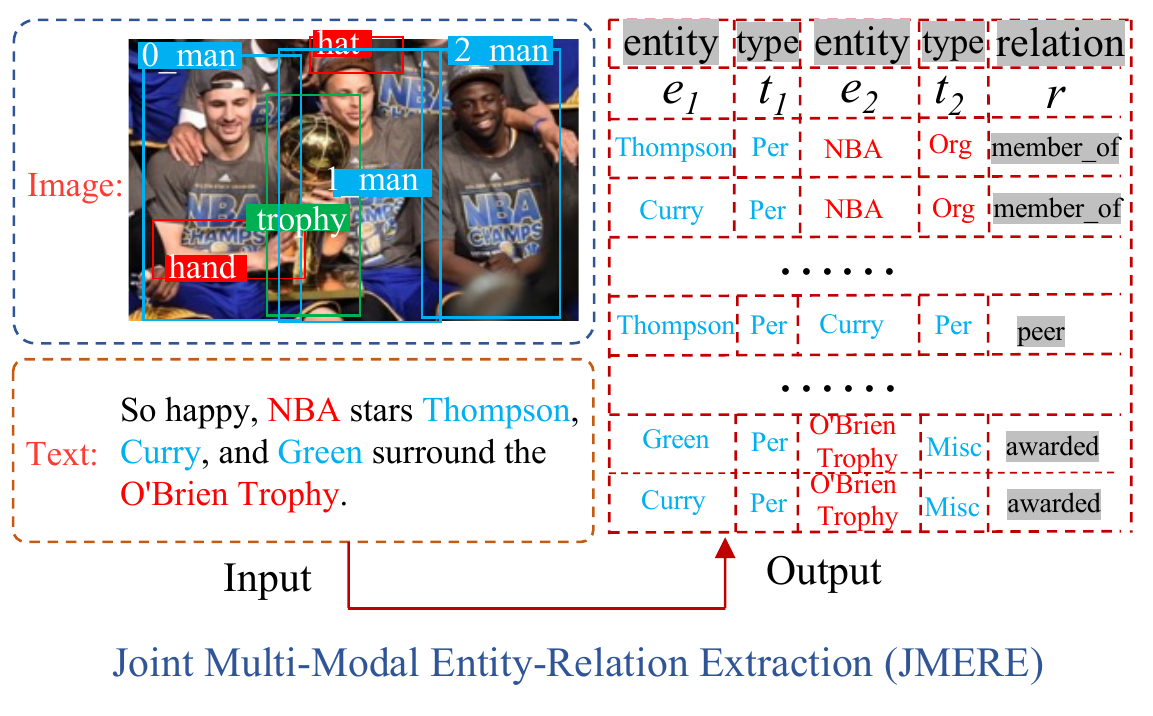}
  \caption{An illustrative example of the joint multimodal entity-relation extraction (JMERE) task, where the Per, Org, and Misc are denoted as the entity types of person, organization, and miscellaneous.}
  \vspace{-0.2cm}
  \label{FIG:1}
\end{figure}

Multimodal named entity recognition (MNER) and multimodal relation extraction (MRE) are two fundamental subtasks for the multimodal knowledge graph construction \cite{liu2019mmkg,Chen2020}, which aims to extend the text-based models by taking images as additional inputs. Previous works usually consider MNER and MRE as two independent tasks \cite{Lu2018,Moon2018,Wu2020OCSGA,Yu2020,Zheng2021,Zhang2021}, which ignore the interaction between these two tasks. Recently, jointing NER and RE as joint entity-relation extraction tasks have attracted much attention in text scenarios, which can exploit the bidirectional interaction between tasks and improve their performance \cite{Wei2020,yuan-etal-2020-graph,Yuan2020}. As shown in Figure~\ref{FIG:1}, if we extract the entity type of (\textbf{Curry}, \textbf{NBA}) is \emph{Per} and \emph{Org}, then their relation should not be \emph{peer}. Otherwise, if we know the relation of entity pair (\textbf{Curry}, \textbf{Thompson}) is the \emph{peer}, then their entity types should be \emph{Per} and \emph{Per}. Thus, the NER can facilitate RE. Meanwhile, RE is also beneficial for NER.

\begin{figure}    
  \centering
    \includegraphics[scale=.58]{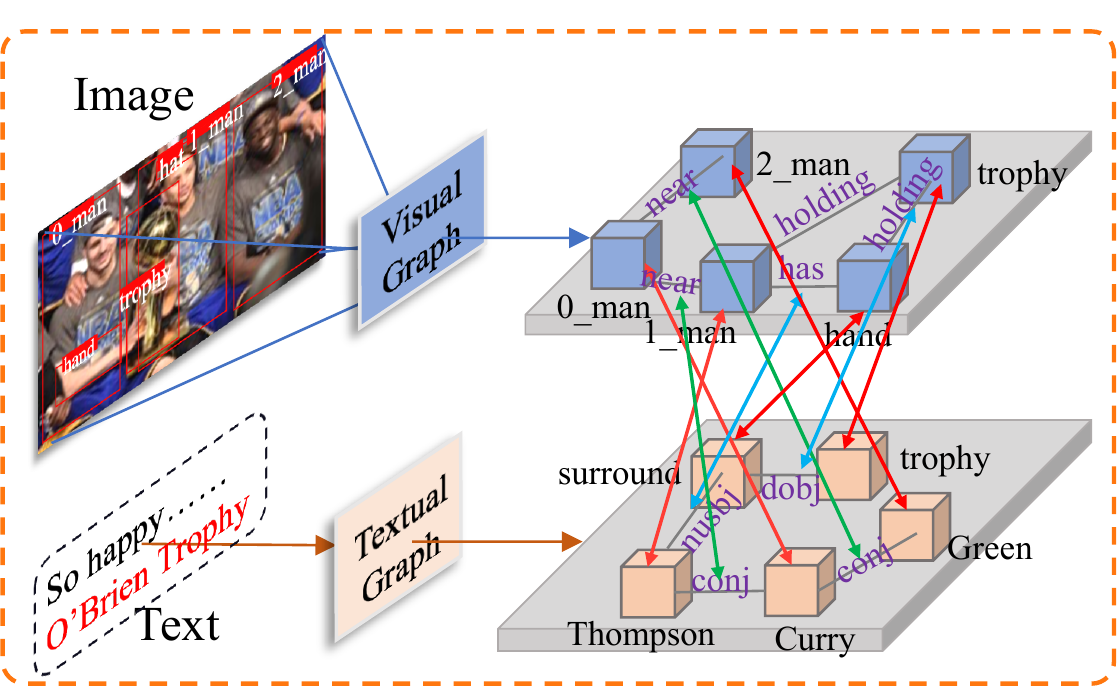}
  \caption{An example for illustrating node (red line) and edge alignment (blue and green lines) in the cross-graph.}
  \vspace{-0.3cm}
  \label{FIG:add}
\end{figure}

However, to the best of our knowledge, joining the MNER and MRE as a joint multimodal entity-relation extraction task (JMERE) has not been studied in the multimodal scenario. Compared with separate tasks, the JMERE task requires extracting different characteristic information from vision. As shown in Figure~\ref{FIG:1}, for the MNER task, if the model can capture the people objects from the image, e.g., outlines of multiple people (blue boxes), it helps to identify the person entity in the text. Meanwhile, the MRE task needs to extract the object-object relationships, e.g., if we know the \emph{holding} is the relationship between \textbf{man\_0} and \textbf{trophy},  we can understand the relation  \emph{awarded} between entities \textbf{Thompson} and \textbf{O'Brien Trophy}. Thus, we consider that the JMERE task should align entities with objects and entity-entity relationships (in text) with object-object relationships (in image). Most recent MNER and MRE studies \cite{Zhang2021,Zheng2021MEGA} align the entities with objects in the visual and textual graphs constructed by the latent relationships of objects and words, as shown by the red lines in Figure \ref{FIG:add}. However, this method only considers node alignment in the cross-graph but ignores edge alignment. As shown in the blue and green lines in Figure \ref{FIG:add}, the edge information in the cross-graph can auxiliary align the nodes and contain clues about the relations between textual entities.

\begin{figure}    
  \centering
    \includegraphics[scale=.64]{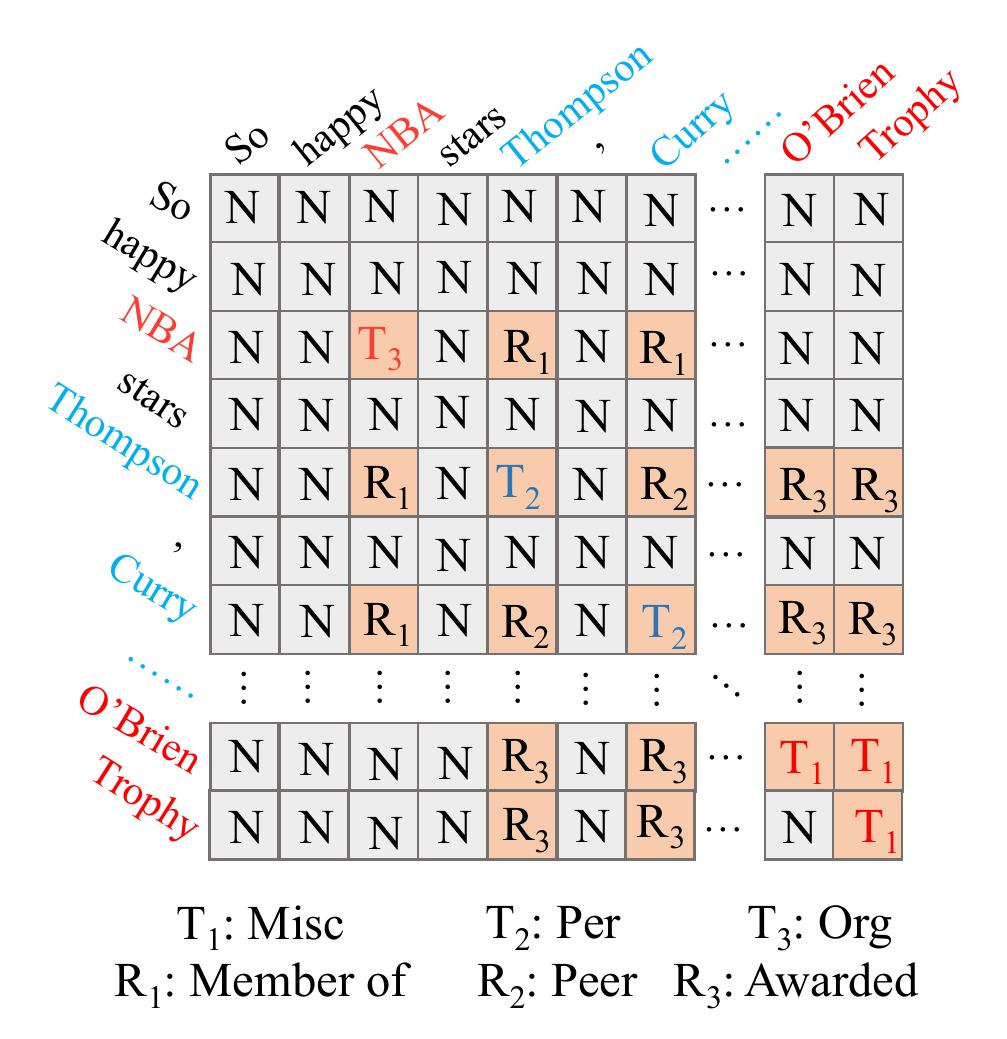}
  \caption{A word-pair relation tagging  example for the JMERE task.}
  \vspace{-0.3cm}
  \label{FIG:2}
\end{figure}

In addition, the pipeline framework method is an intuitive way to solve the JMERE task. It extracts the entities using MNER method and then classifies their relations by MRE method. However, the pipeline framework only benefits MRE through the results of MNER and suffers from error propagation \cite{Ju2021}. As shown in Figure~\ref{FIG:1},  if the MNER extracts the entity type of (\textbf{Curry}, \textbf{NBA Stars}) is \emph{Per} and \emph{Misc}, the result should be incorrect. Inspired by the grid tagging scheme in the aspect-based sentiment triplet extraction task \cite{Wu2020}, we first adopt a word-pair ($w_i$,$w_j$) classification scheme for the JMERE task, namely the word-pair relation tagging. This scheme simultaneously trains MNER and MRE tasks to exploit the interaction between them and avoid the error propagation caused by the pipeline framework. As shown in Figure~\ref{FIG:2}, the word-pair relation tagging of word pairs (\textbf{Curry}, \textbf{Curry}) and (\textbf{Thompson}, \textbf{Curry}) is \emph{Per} and \emph{Peer}, which denotes \textbf{Curry} belongs to a person and the \emph{peer} denotes this the relation between \textbf{Curry} and \textbf{Thompson}.

\begin{figure*}    
  \centering
    \includegraphics[scale=.55]{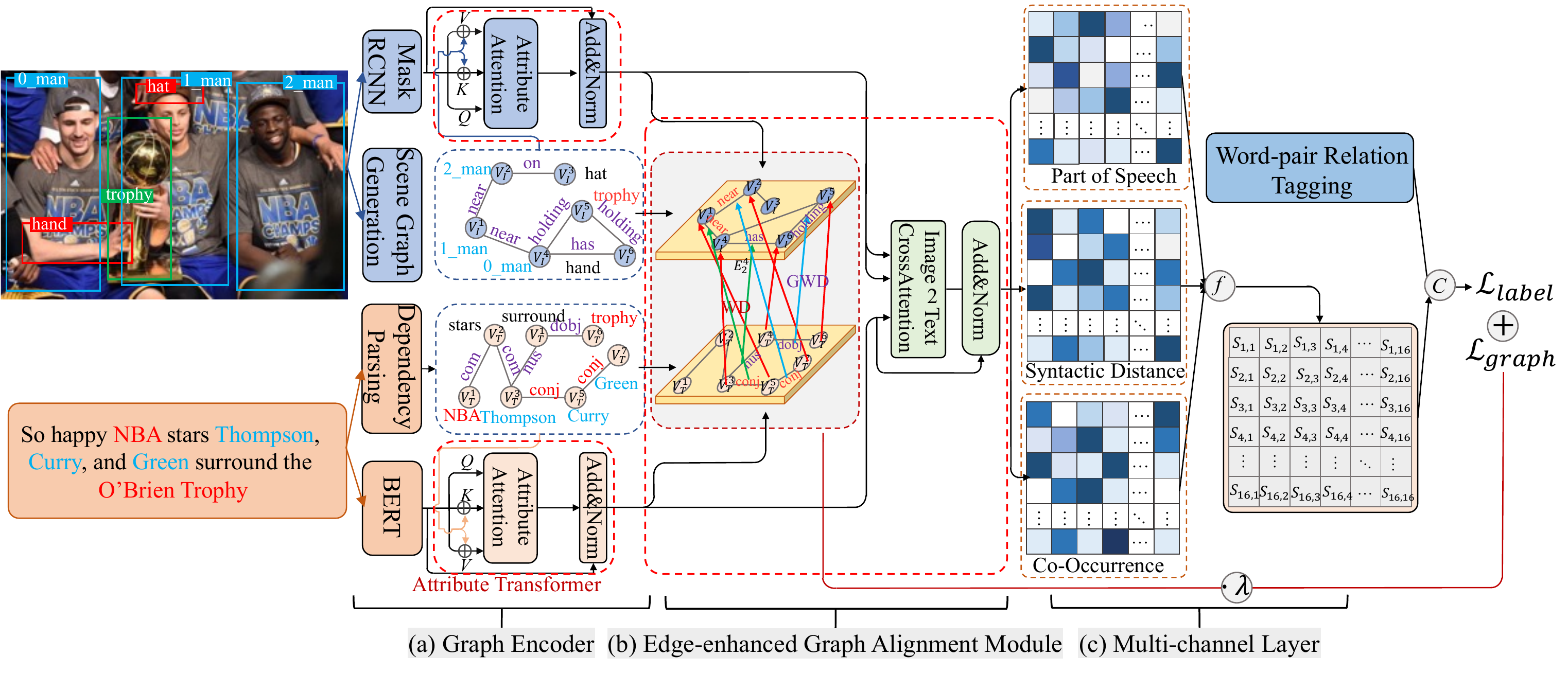}
  \caption{The overall framework of EEGA for JMERE, where com and nus denote the syntactic dependency relationships compound and nsubj.}
  \vspace{-0.3cm}
  \label{FIG:3}
\end{figure*}

To address the above challenges, we propose an edge-enhanced graph alignment network (EEGA) and word-pair relation tagging to enhance JMERE by simultaneously aligning objects with entities (e.g., \textbf{0\_man} with \textbf{Curry} and \textbf{trophy} with \textbf{trophy}) and object-object relationships with entity-entity relationships (e.g., \emph{near} with \emph{conj} and \emph{holding} with \emph{dobj}) in the cross-graph. The overall framework of EEGA is shown in Figure~\ref{FIG:3}. Specifically, we use a graph encoder layer to construct the textual and visual graphs from the input text-image using pre-trained models. Then, we propose an edge-enhanced graph alignment module with Wasserstein distance to align the nodes and edges in the cross-graph. Meanwhile, the module can leverage the edge information to auxiliary alignment between objects and entities and find the correlations between entity-entity relationships and  object-object relationships. Finally, we design a multi-channel layer by mining word-word relationships from different perspectives to obtain the final word pair representations. 

Our main contributions can be summarized as follows:
\begin{itemize}
\item We are the first to propose the joint multimodal entity relation extraction (JMERE) task to handle the multimodal NER and RE tasks. Meanwhile, we design a word-pair relation tagging for JMERE. This scheme can exploit the bidirectional interaction between MNER and MRE and avoid the error propagation caused by the pipeline framework.
\item We propose an edge-enhanced graph alignment network (EEGA) to enhance the JMERE task by aligning nodes and edges simultaneously in the cross-graph. Compared with previous methods, the EEGA can leverage the edge information to auxiliary alignment between objects and entities and find the correlations between entity-entity relationships and object-object relationships.
\item  We conduct extensive experiments on the collected JMERE dataset, and the experimental results demonstrate the eﬀectiveness of our proposed model.
\end{itemize}

\section{Related works}
The crucial components of the knowledge graphs construction task \cite{Chen2020,chen-etal-2022-good,10.1145/3477495.3531992}, named entity recognition (NER) and relation extraction (RE), have attracted much attention from researchers \cite{Vashishth2018,Wen2020,Li2020,ren-etal-2020-two, Nasa2021,zhao2021asking}. Previous researches have mainly focused on a single modality. With the increasing popularity of multimodal data on social platforms, some studies have begun to focus on multimodal NER (MNER) and multimodal RE (MRE), which aim to consider the image as a supplement to text and better recognize the entities and their relations. According to the object of image-text alignment, the current methods of MNER and MRE can be divided into image alignment methods, object alignment methods, and node alignment methods.

\subsection{Image Alignment Methods} Previous studies usually used RNN (Recurrent Neural Networks) to encode text and CNN (Convolution Neural Network) to encode image as a  vector. Then, designing an implicit interaction module to model the information between modalities for the MNER task \cite{Zhang2018,Lu2018,Moon2018}. For example, \citet{Zhang2018} constructed an MNER dataset and proposed a baseline model based on a bi-directional long-term memory network using an attention mechanism to align the image representation with text. However, encoding image as a vector cannot benefit extract different type entities, e.g., \textbf{Curry} (\emph{Per}) and \textbf{O'Brien Trophy} (\emph{Misc}).

\subsection{Object Alignment Methods} To address limitation of image alignment methods, the previous models extracted different visual objects using Mask-RNN or Fast-RNN \cite{he2017mask} and aligned visual objects with the text representation \cite{Wu2020OCSGA,Yu2020,Zheng2021,Zhang2021,Xu2022}. \citet{Wu2020OCSGA} proposed an interactive attention structure to align text with visual objects. In addition, \citet{Zheng2021} designed a gated bilinear attention network \cite{Kim2018} with an adversarial strategy to better extract the fine-grained objects from the image. However, the object alignment method does not consider the relations of entity-entity and object-object, and the model will ineffectively match overlapping visual objects with textual entities.  For example, the \textbf{trophy} of text may align with the \textbf{1\_man}, since the \textbf{1\_man} contains the region of \textbf{trophy}. 

\subsection{Node Alignment Methods} To address the above limitation, the most current researches align the entities with objects in the visual and textual graphs constructed
by the latent relationships of objects and words \cite{Zhang2021,zhang2021ma,Zheng2021MEGA}. \citet{Zhang2021} proposed a graph-based multimodal fusion model based on the syntactic dependency text graph and full connection visual graph to exploit the fine-grained semantic alignment different modalities. In the MRE task \cite{Zheng2021MNRE}, \citet{Zheng2021MEGA} designed a graph alignment module to align nodes in textual graph and visual graphs. However, the node alignment methods only consider the nodes in the cross-graph and ignore the edge information. The edge information in the cross-graph can effectively improve the matching precision of nodes and contain some clues about the classifying relation between entities.
\section{Task Definition and Word-pair Relation Tagging}
\subsection{Task Definition}
The joint multimodal entity-relation extraction task is defined: given an input text $w=\left\{ w_{1},w_{2},\cdots ,w_{n}\right\}$ with a corresponding image \emph{I}, the model is required to extract a set of quintuples $y=\left\{\left(e_{1}, t_{1}, e_{2}, t_{2}, r\right)_{c}\right\}_{c=1}^{C}$, where ${\left(e_{1}, t_{1}, e_{2}, t_{2}, r\right)_{c}}$ represents the \emph{c}-th quintuple, consisting of two entities $e_1$ and $e_2$ with the corresponding entity types $t_1$ and $t_2$, where ${e_1}\neq {e_2}$. Furthermore, $r$ represents the relation between the entities $e_1$ and $e_2$.  Figure~\ref{FIG:1} gives an example to better understand the JMERE task, which aims to extract all quintuples, e.g., (\textbf{Thompson}, \emph{Per}, \textbf{NBA}, \emph{Org}, \emph{Member\_of}), where \emph{Per} and \emph{Org} indicate the entity types of \textbf{Thompson} and \textbf{NBA}, and \emph{Member\_of} denotes their relation type. 

\subsection{Word-pair Relation Tagging}
Inspired by the grid tagging scheme in aspect-based sentiment triplet extraction \cite{Wu2020}, we design a word-pair relation tagging to extract all elements of JMERE in one step. By word-pair relation tagging, the JMERE task is converted into extracting the relations $Y$ between each word-pair $(w_{i},w_{j})$ and avoid the error propagation caused by the pipeline framework. These relations can be explained below, and we also give an example in Figure~\ref{FIG:2} to better understand the word-pair relation tagging. 
\begin{itemize}
\item \emph{N} indicates that the word-pair does not have any relation.

\item \textbf{$T$} indicates that the word-pair belongs to an entity type, which is contained 4 defined types in the previous work \cite{Zheng2021}

\item \textbf{$R$} indicates that the word-pair belongs to defined relation \cite{Zheng2021MEGA} and each word is an entity.
\end{itemize}
\section{Edge-enhanced Graph Alignment Network}
Figure~\ref{FIG:3} shows the overall architecture of the proposed model, consisting of three parts: a graph encoder, an edge-enhanced graph alignment module, and a multi-linguistic channel layer. The graph encoder layer uses the pre-trained models to construct the textual and visual graphs from the input. To enhance the ability to capture edge information, we do not directly send textual and visual representations to the next module but an attribute transformer. Then, to match the objects with entities more precisely and capture the entity-entity relation clues from the visual graph, we use a cross-graph optimal transport method \cite{chen2020graph} with Wasserstein distance and Gromov-Wasserstein distance to simultaneously align the nodes and edges in the cross-graph. Finally, we propose a multi-channel layer that uses a weighted graph convolution network (W-GCN) to mine the latent relationships for word pairs from multi-perspectives. A detailed description of each component is provided below. The code of the manuscript will publish in the final version.
\subsection{Graph Encoder}

\subsubsection{Textual Graph.} Formally, we first use the dependency parse toolkit\footnote{https://spacy.io/models.} to construct the textual graph. As shown in Figure~\ref{FIG:3} (a), after parsing, the given sentence is converted into a textual graph ${G_{T} = \left\{V_{T},E_{T}\right\}}$, where $V_{T} \in\mathbb{R}^{{n}}$ and ${E_{T} \in\mathbb{R}^{{n \times n}}}$ denote syntactic dependency nodes and edges, respectively, and $G_{T}$ is an un-directed self-loop graph. Meanwhile, we use $A_T \in\mathbb{R}^{{n \times n}}$ to denote the adjacent mask matrix for textual graph, where $A_T^{i,j}\in\{0,1\}$ indicate whether there is an edge between $w_i$ and $w_j$ or not. Furthermore, the nodes $V_{T}$ are fed into BERT to obtain $X_{T} \in\mathbb{R}^{n \times d_{T}}$. Meanwhile, an edge transition matrix is used to map the edge type $E_{T}^{i,j}$ into a trainable vector and obtained edge trainable matrix $Z_{T} \in\mathbb{R}^{n \times n\times d_{zT}}$, 
\begin{equation}
\label{eq:1}
X_{T}=BERT\left(w\right)
\end{equation}
where $BERT$ denotes the BERT as the text encoder and $d_{T}$ is the dimension of the BERT output.

\subsubsection{Visual Graph.} In previous multimodal tasks, the objects were considered the semantic information of images. As shown in Figure~\ref{FIG:3} (a), we convert the image into a visual graph $G_{I} = \left\{V_{I},E_{I}\right\}$ by using the scene graph generation model \cite{Tang2020} (the Mask-RCNN used as the backbone). Furthermore, we only consider the top \emph{k} salient objects with the highest object classification scores as the valid visual objects and exploit the adequate visual information while ignoring the irrelevant ones. Thus, the final node represents $V_{I}\in \mathbb{R}^{k}$  salient objects detected by Mask-RCNN and $E_{I} \in\mathbb{R}^{k \times k}$ denotes the visual relationship set, such as position relationships (e.g., \emph{near} and \emph{in front of}) and affiliation relationships (e.g., \emph{holding} and \emph{wearing}). We use $A_I \in\mathbb{R}^{k \times k}$ to denote the adjacent matrix of the visual graph. Thus, the final node vectors in visual graph $X_{I} \in\mathbb{R}^{k \times d_{I}}$ are defined as, 
\begin{equation}
\label{eq:2}
X_{I}={Mask\mbox{-}RCNN}\left(I\right)
\end{equation}
where $d_{I}$ is the hidden dimension of Mask-RCNN and the final edge vectors $Z_{I} \in\mathbb{R}^{k \times k\times d_{zI}}$ are obtained in the same way as the textual graph.

\subsubsection{Attribute Transformer.}  we propose the attribute attention (\emph{At-Att}) by incorporating the edge types into keys and values in the self-attention of the transformer as the attribute transformer, which can update the node state in the inter-model while effectively incorporating relationship edges in the cross-graph (e.g., \emph{nsubj} and \emph{acomp} in the textual graph, and \emph{holding} and \emph{wearing} in the visual graph). 

Since the visual and textual graphs are two semantic units containing information in different modalities, we model them using similar operations but with different parameters. Thus, the hidden representation of  the \emph{i}-th token $\widetilde{H}_{T}^{i} \in \mathbb{R}^{n \times d_{T}}$ in text modalities is defined as,
\begin{eqnarray}
\label{eq:3}
\widetilde{H}_{T}^{i} &=& \mathop{At\mbox{-}Att}\left(X_{T}^{i}, Z_{T}^{i}, A_{T}^{i}\right) \nonumber \\ 
  {}&=& \mathop{\rm Softmax} \left(A_{T}^{i} \cdot \frac{Q_{T}^{i} ({K_{T}^{i}})^\top}{\sqrt{d_{T}}}\right){V_{T}^{i}} \end{eqnarray}
where ${A_{T}^{i}} \in\mathbb{R}{^{1 \times n}}$ is the adjacency mask set of the \emph{i}-th node, ${Q_{T}^{i}} \in \mathbb{R}{^{1 \times d_T}}$ , ${K_{T}^{i}} \in\mathbb{R}{^{n \times d_T}}$, and ${V_{T}^{i}} \in\mathbb{R}{^{n \times d_T}}$ are matrices that package the queries, keys, and values for the \emph{i}-th word in text correspondingly, which are defined as, 
\begin{eqnarray}\label{eq:4}
{Q_{T}^{i}} &=& {W_Q}X_{T}^{i} \nonumber \\ 
{K_{T}^{i}} &=& {W_K}X_{T} + {W_z}{Z_{T}^{i}} \\
{V_{T}^{i}} &=& {W_V}X_{T} + {W_r}{Z_{T}^{i}} \nonumber
\end{eqnarray}
The other operations of the attribute transformer are consistent with the vanilla transformer: $\widetilde{H}_{T}$ is added to $X_T$ using a feed-forward network (FFN) and layer normalization (LayerNorm) to obtain the text representation ${H}_{T} \in \mathbb{R}^{n \times d_{T}}$.

We use similar operations to obtain the visual representation. In particular, we use a variable-dimensional FFN to match the dimension of object ${H}_{I}$ and token ${H}_{T} $. Thus, the image representation of graph encoder can be denoted as ${H}_{I} \in \mathbb{R}^{k \times d_{T}}$.

\subsection{Edge-enhanced Graph Alignment Module}
Given the textual graph $G_{T} = \left\{H_{T},Z_{T}\right\}$, and the visual graph $G_{I} = \left\{H_{I},Z_{I}\right\}$. We aim to align the nodes and edges of the cross-graphs simultaneously and to transfer the matched semantic information from objects into entities. Formally, an optimal transport method \cite{chen2020graph} is first used to match the nodes and edges in the cross-graph. Furthermore, we use image2text attention to transfer matched semantic information from the visual object to text modality and obtain the refined textual representation.
\begin{figure}    
  \centering
    \includegraphics[scale=.67]{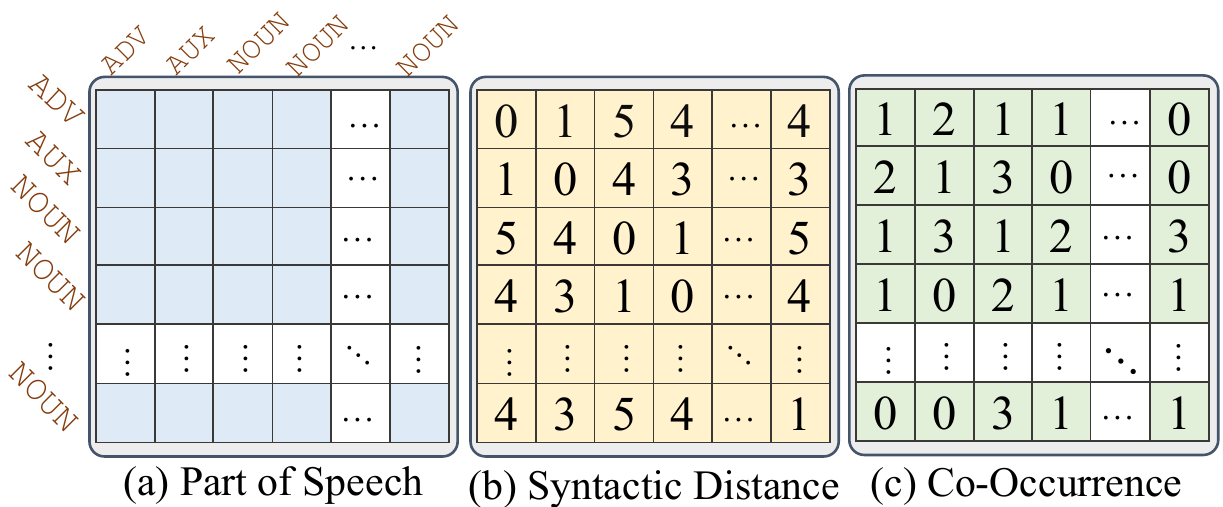}
  \caption{The multi-channel matrices of the given sentence.}
  \vspace{-0.3cm}
  \label{FIG:4}
\end{figure}
\subsubsection{Edge-enhanced Graph Optimal Transport.} To explicitly encourage simultaneously aligning the nodes and edges in the cross-graph, we apply the optimal transport method initially proposed in transfer learning. As illustrated in Figure~\ref{FIG:3} (b), unlike the original optimal transport method considering text and image as full-connected graphs, we only consider the nodes and edges of having adjacency relationships in the cross-graph.
Particularly, two types of distance are adopted for cross-graph matching: (1) Wasserstein Distance (WD) \cite{peyre2019computational} for node matching (the red lines); (2) Gromov-Wasserstein Distance (GWD) \cite{peyre2016gromov} for edge matching (the blue and green lines). Formally, the $D_{wd}({H}_{I}, {H}_{T})$ is measured the optimal transport distance to match the nodes ${H}_{I}$ to ${H}_{T}$, which is defined as:
\begin{eqnarray}\label{eq:5}
{D_{wd}}({H_I},{H_T}) &=& \min \sum\limits_{i = 1}^k {\sum\limits_{j = 1}^n {{{T}_{i,j}} \cdot c(H_I^i,H_T^j)} }
\end{eqnarray}
where $c(H_I^i,H_T^i)$ denotes the cosine distance between $X_I^i$ to $X_T^j$, which is defined as $c(H_I^i,H_T^j) = 1 - \frac{{{{(H_I^i)}^T}H_T^j}}{{\left\| {H_I^i} \right\|\left\| {H_T^j} \right\|}}$. The matrix $T$ is the transport information flow, where $T_{i,j}$ represents the amount of cost shifted from node $X_I^i$ to $X_T^j$. 

Then, we use the Gromov-Wasserstein distance \cite{peyre2016gromov} to measure the similarity scores ${D_{gwd}}$ of the edge in the cross-graph by calculating distances between node-pairs, which is defined as,

\begin{eqnarray}\label{eq:6}
&& {D_{gwd}}({H_I},{H_T},H_I^{{'}},H_T^{{'}}) = \nonumber \\
&&\min \sum\limits_{i = 1}^k {\sum\limits_{j = 1}^n {{{\hat T}_{i,j}}{{\hat T}_{{i'},{j'}}}} } \cdot  {L}\left(H_I^{i},H_I^{{i'}}, H_T^j, H_T^{{j'}} \right)
\end{eqnarray}
where $H_T^{{j'}}$ and $H_I^{{i'}}$ are adjacent nodes sets in the textual and visual graphs of $H_T^{j}$ and $H_I^{i}$ respectively, and $L(\cdot )$ is considered as the distance cost of the cross-graph edges $(H_I^{i},H_I^{{i'}})$ to $(H_T^j, H_T^{{j'}}$), i.e. $L(H_I^i,H_I^{i'},H_T^j,H_T^{j'}) = \left\| {c(H_I^i,H_I^{i'}) - c(H_T^j,H_T^{j'})} \right\|$. The learned matrix ${\hat T}$ now denotes a transport plan that aids in aligning edges in the cross-graph.

 We use a unified solver and use the Sinkhorn algorithm \cite{NIPS2013_af21d0c9} with an entropic-regularizer \cite{benamou2015iterative} to iteratively optimize costs ${D_{wd}}$ and ${D_{gwd}}$. Thus, the object loss function of optimizing the cross-graph is,
\begin{eqnarray}\label{eq:7}
\mathcal{L}_{graph} = \alpha  \cdot {D_{wd}}({H_I},{H_T}) + (1 -\alpha )\cdot \nonumber \\  
{D_{gwd}}({H_I},H_{I}^{'},{H_T},H_{T}^{'}) 
\end{eqnarray}
where $\alpha$ is the hyper-parameter for balancing the importance of costs. Then, we use image2text attention to effectively transform the visual semantic information into the textual representation, which is denoted as, 
\begin{eqnarray}\label{eq:8}
\widetilde{O}= {ATT_{cross}}({{H}_T},{{H}_I},{H_I})\
\end{eqnarray}
where ${ATT_{cross}}$ denotes the cross-modal multihead attention \cite{101145}. Then, the $\widetilde{O}$ is added with ${H}_T$ and sends a layer normalization to obtain the final contextual representation ${O}$. 
\begin{table*}[]
\centering
\begin{tabular}{|c|c|c|c|c|c|c|c|}
\hline
\multicolumn{2}{|c|}{\multirow{2}{*}{Methods}}                                                    & \multicolumn{3}{c|}{JMERE} & \multicolumn{3}{c|}{\#MNER} \\
\cline{3-8}
\multicolumn{2}{|c|}{}                                                                            & \#P    & \#R    & F1    & \#P     & \#R     & F1     \\
\hline
\multirow{4}{*}{Pipeline Methods}                               & AdapCoAtt+MEGA & 48.44  & 47.06  & 47.74 & 74.32   & 72.11   & 73.20  \\
& OCSGA+MEGA     & 48.21  & 47.99  & 48.10 & 75.27   & 72.32   & 73.77  \\
& AGBAN+MEGA     & 47.87  & 48.28  & 48.57 & 74.78   & 73.69   & 74.23  \\

& UMGF+MEGA      & 49.28  & 50.76  & 50.01 & 75.02   & 76.77   & 75.88  \\
\hline
\multirow{7}{*}{\begin{tabular}[c]{@{}c@{}}Word-pair Relation \\ Tagging  Methods \end{tabular}} & AdapCoAtt$^*$    & 50.22  & 47.67  & 48.91 & 77.32   & 73.28   & 75.25  \\
 & OCSGA$^*$        & 52.11  & 47.41  & 49.64 & 77.13   & 75.03   & 76.07  \\
& AGBAN$^*$        & 51.07  & 48.89  & 49.95 & 76.57   & 75.82   & 76.19  \\
& UMGF$^*$         & 52.76  & 50.22  & 51.45 & 77.51   & 76.01   & 76.75  \\
& MEGA$^*$         & 55.08  & 51.40  & 53.18 & 77.78   & 76.67   & \underline{77.22}  \\
& MAF$^*$         & 52.56  & \textbf{54.73}  & \underline{53.62} & 76.07  & 77.57  & 76.81  \\
\cline{2-8}
& EEGA(\emph{ours})     & \textbf{58.26}\dag  & \underline{52.61}  & \textbf{55.29}\dag & \textbf{78.27}   & \textbf{78.91}\dag   & \textbf{78.59}\dag \\
\hline
\end{tabular}
\caption{The experiment results on the JMERE task (\%) and the \#MNER denote the MNER results computed by the JMERE results. AGBAN$^*$ means using the word-pair relation tagging in the AGBAN model. The marker {\dag} refers to significant test $p-value<0.05$. The best result is in bold and \#P, \#R, and F1 denote the precision, recall, and F1-score. }
\label{table:2}
\end{table*}
\subsection{Multi-channel Layer}
In this subsection, we aim to mine the different dependency features between $w_i$ and $w_j$ to help detect relations between them. As shown in Figures \ref{FIG:4}: (a) we consider that part of speech (Pos) can provide lexical information for word pairs. For example, the Pos of most entities belong to the \texttt{NOUN} and \texttt{PEROPN}, e.g., \textbf{NBA}, \textbf{Curry}, and \textbf{Thompson}; (b) encoding the syntactic distance (Sd) between word-pair can improve the ability of the model to capture long-range syntactic information; (c) the word co-occurrences matrix (Co) can provide corpus-level information between word pairs. For example, \textbf{Curry} and \textbf{NBA} appear some times in the corpus. The details about constructing each feature matrix are added in Appendix-A.

After data preprocessing, three feature matrices are obtained, ${M^l} \in {\mathbb{R}^{n \times n}},l \in \{ Pos,Sd,Co\}$. We propose a W-GCN module to model each matrix, obtaining each channel representation. Each matrix ${M^l}$ firstly sends an embedding layer yielding a trainable representation ${R^l}\in {\mathbb{R}^{n \times n\times d_l}}$ and $d_l$ is the dimension of representation. The calculation W-GCN process of \emph{i}-th word in \emph{l}-th matrix is shown as, 
\begin{eqnarray}\label{eq:9}
\begin{array}{l}
S_i^l = {\rm{W\mbox{-}GCN}}_l(R_i{{^l}},{O})\\
 = {\rm{Softmax}}({\rm{ReLU}}(W_l^{{r_1}}R{_i^{l^T}} + {b_l}))\cdot(W_l^{{r_2}}(O))
\end{array}
\end{eqnarray}
where $R_i^l\in {\mathbb{R}^{n\times d_l}}$ is the \emph{i}-th word in \emph{l}-th linguistic matrix  and $W_l^{{r_1}}{\in\mathbb{R}^{1\times d_l}}$ and $W_l^{{r_2}}{\in\mathbb{R}^{d_T\times d_T}}$ are shared weights used to perform a linear layer to learn linguistic features and representational abilities. We combine the representations and send them to the MLP (\textbf{M}ulti\textbf{L}ayer \textbf{P}erception) layer for obtaining the final word representation, 
\begin{eqnarray}\label{eq:10}
{S_i} = {\rm{MLP}}[S_i^{Pos};S_i^{Sd};S_i^{Co}]
\end{eqnarray}
where $S_i \in \mathbb{R}^{d_T} $ is the \emph{i}-th word representation. Thus, the output representation of the multi-channel layer is denoted as $S = [{S_1},{S_2} \cdots ,{S_n}]$. Finally, we concatenate the enhanced representations of $S_i$ and $S_j$ to represent the word-pair $({w_i}, {w_j})$, i.e., $r_{i,j}=[S_i; S_j]$. Then, send the $r_{i,j}$ to a  linear prediction layer and obtain the probability distribution,
\begin{eqnarray}\label{eq:11}
{p_{i,j}} = {\rm{Softmax(}}{W_p}{r_{i,j}} + {b_p}{\rm{)}}
\end{eqnarray}
where $W_p \in \mathbb{R}^{d_y \times 2d_T}$ and $b_p\in \mathbb{R}^{d_y}$ are trainable parameters and $d_y$ is the number of tags. Then, we used the cross-entropy error to measure the ground truth distribution $\mathcal{Y}$ and predicted tagging distribution,
\begin{eqnarray}\label{eq:12}
\mathcal{L}_{main}(\theta ) = {-}\sum\limits_{s = 1}^S {\sum\limits_{i = 1}^n {\sum\limits_{j = 1}^n {\mathcal{Y}_{i,j}^s\log (p_{i,j}^s\theta )} } }
\end{eqnarray}
where \emph{S} and $\theta$ denote the number of training samples and all trainable parameters, respectively.
\subsection{Join Training}
The final objective is a combination of the main task and optimizing the cross-graph as follows,
\begin{eqnarray}\label{eq:13}
\mathcal{L}={\mathcal{L}_{main}} + \lambda  \cdot {\mathcal{L}_{graph}}
\end{eqnarray}
where $\lambda $ are trade off hyper-parameters to control the contribution of optimizing the cross-graph.
\begin{table}[]
\centering
\begin{tabular}{|c|c|c|c|}
\hline
Methods                     & \#P   & \#R   & F1    \\
\hline
EEGA(\emph{All})                  & 58.26 & 52.61 & 55.29 \\
\hline
\emph{w/o edge-enhanced}            &  51.85     &50.31       & 51.07      \\
\hline
\emph{w/o attribute transformer}   
&    55.69    & 51.07     &     53.28 
    \\
\hline
\emph{w/o multi-channel}         &    55.48   &  52.13     & 53.75     \\
\hline
\end{tabular}
\caption{Results of ablation study for the JMERE task.}
\label{table:3}
\end{table}
\begin{figure*}   
  \centering
    \includegraphics[scale=.50]{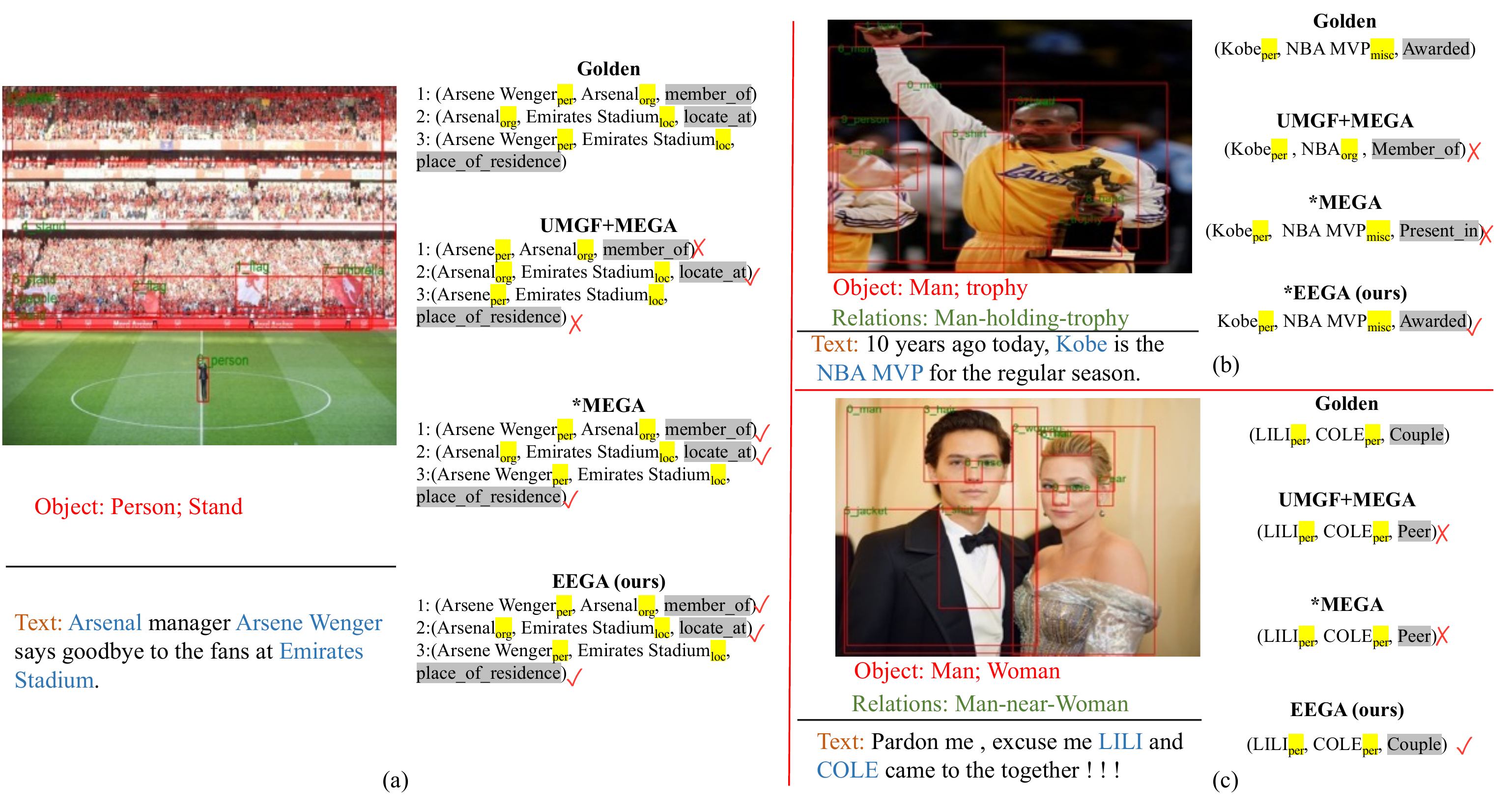}
  \caption{Three cases of the predictions by UMGF+MEGA,*MEGA, and EEGA.}
  \vspace{-0.3cm}
  \label{FIG:6}
\end{figure*}
\section{Experiments}
Comparative experiments were conducted to evaluate and compare the performance of the EEGA method against several prior works. Furthermore, more detailed experiments (e.g., datasets, setting, and parameter sensitivity) are presented in Appendix B-D.

\subsection{Comparative Results}
\noindent \textbf{Compared Methods.} We summarize the MNER and MRE studies and combine the state-of-the-art methods as our strong JMERE baselines, as shown in Table~\ref{table:2}. They include AdapCoAtt \cite{Zhang2018}, OCSGA \cite{Wu2020OCSGA}, AGBAN \cite{Zhang2021}, and MAF \cite{Xu2022}  for extraction of the entity and the corresponding type,  and UMGF \cite{Zheng2021MEGA} for relation extraction of entities. In addition, we apply the word-pair relation tagging to the above baseline models to investigate the effectiveness of the word-pair relation tagging and the proposed method, e.g., AdapCoAtt$^*$, OCSGA$^*$, and MEGA$^*$.

\noindent \textbf{Overall Results.} Observing pipeline methods, we find that the UMGF+MEGA performs better than other pipeline methods, which shows that aligning the nodes in the cross-graph can benefit matching the entities with objects. The word-pair relation tagging methods outperform the pipeline methods in JMERE and \#MNER, such as OCSGA, AGBAN, and UMGF, showing that the word-pair relation tagging can improve performance by leveraging task relationships and reducing error propagation issues caused by the pipeline framework.

Furthermore, the EEGA surpasses all baselines. Compared with the best results of existing baselines, EEGA still achieves absolute F1-score increases of 1.67\% and 1.37\% on JMERE and \#MNER. The experimental results strongly prove that simultaneously aligning the nodes and edges in the cross-graph can effectively improve the precision of matching the objects with entities and capture more relationships between entities. In addition, the proposed attribute transformer enhances the ability to mine relations between nodes by incorporating edge information into the key and value in the transformer. Meanwhile, the multi-channel layer can take linguistic relations between word pairs to refine the final representation and improve prediction performance.

\noindent \textbf{Ablation Study.} To investigate the effectiveness of different components in EEGA, edge-enhanced graph optimal transport (edge-enhanced), attribute transformer, and multi-channel layer, we conduct an ablation study for the JMERE task in Table~\ref{table:3}. \emph{W/o attribute transformer} means that a vanilla transformer replaces the attribute transformer. The F1-score dropped 2.01\%, indicating that integrating the edge information into the key and value in the transformer can enhance the ability to capture the relations between nodes and benefit the edge alignment in the cross-graph. The performance is decreased after removing the multi-channel layer (\emph{w/o multi-channel}), indicating the multi-channel layer can mine relationships of word pairs from different perspectives and refine the final representation.

\emph{W/o edge-enhanced} means removing the edge-enhanced graph optimal transport from EEGA. The performance of the model is highly degraded after removing the edge-enhanced, showing that simultaneously aligning nodes and edges in the cross-graph can be beneficial for matching the visual objects with textual entities more precisely and finding the entity
classification clues from the relationship between objects.
\begin{figure}[] %
  \centering  
  \subfigcapskip=-5pt 
  \subfigure[$\alpha=1$.]{
    \label{level.sub.1}
    \includegraphics[width=0.45\linewidth]{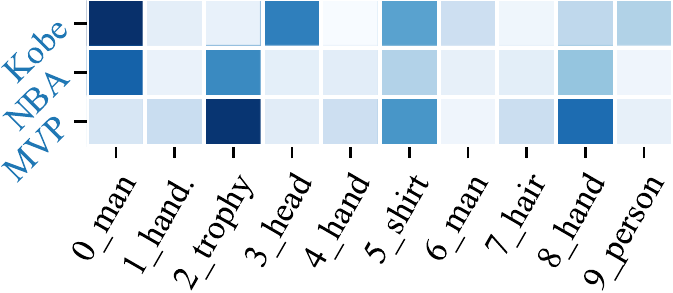}}
  \subfigure[$\alpha=0.4$.]{
    \label{level.sub.2}
\includegraphics[width=0.45\linewidth]{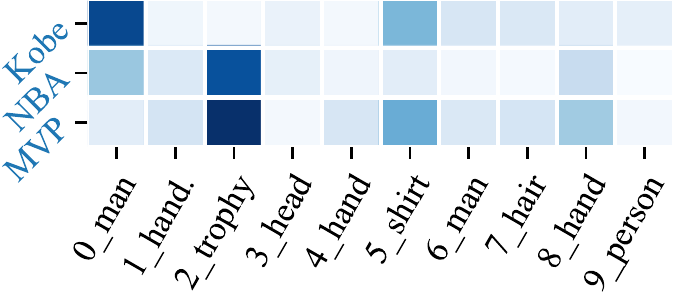}}
  \caption{A comparison of Attention visualization on entity-object pair from $\alpha=1$ (only node alignment) and $\alpha=0.4$ (best
performance setting) in Eq. (\ref{eq:7}) }
    \vspace{-0.3cm}
  \label{FIG:visualization}
\end{figure}
\subsection{Case Study}
To understand the effectiveness of our proposed model, Figure~\ref{FIG:6} presents three examples with the predicted results. Meanwhile, the important objects and relations are detected from images. In example (a), only the pipeline-based model UMGF+MEGA extracts incorrectly since the pipeline model easily suffers from error propagation, i.e., the extraction entity \textbf{Arsene} by UMGF is incomplete, and the final model UMGF+MEGA extract the incorrect quintuple. In example (b), UMGF+MEGA imprecisely extracts \textbf{NBA} as an entity, and *MEGA incorrectly predicts the relationship between \textbf{Kobe} and \textbf{NBA MVP} as the \emph{Present\_in}. In example (c), the situation is similar to example (b). Since lacking effective ways to map the semantic relationship of objects \emph{Man-near-Woman} to entities (\emph{LILI-COLE}), the UMGF+MEGA and *MEGA incorrectly predicts relations \emph{peer} between entities.

For these three examples, the proposed EEGA makes accurate judgments. Benefiting from the edge-enhanced graph optimal transport module,  the EEGA can align the nodes and edges in the cross-graph to match the entities with objects more precisely. Meanwhile, the EEGA also effectively captures the relation clue from the visual graph to the textual graph shown in examples (b) and (c). In addition, the attribute transformer and multi-channel layer can further enhance the ability to model the relationships of objects and word pairs.
\subsection{Visualization Analysis} In this section, we visualize the example (b) of Figure~\ref{FIG:6} at $\alpha=1$ and $\alpha=0.4$ to test whether our edge alignment strategy helps to learn fine-grained entity-object matching. As shown in Figure~\ref{FIG:visualization} (a), when only node alignment means {$\alpha=1$}, since the proposed model lacks the edge constraints, the attention weight is relatively scattered and affects the precision of matching entities with objects. Particularly, the model easily classifies the type of entity \textbf{NBA} as \emph{Per}. Meanwhile, as shown in the \textbf{NBA} and \textbf{Kobe} in Figure \ref{FIG:visualization} (b), benefiting from the edge alignment can find the mapping between the object-object relationships and entity-entity relationship; the EEGA effectively reduces the ambiguity and match objects with entities more precisely.

\section{Conclusion}
In this paper,  we are the first to propose a joint multimodal entity relation extraction (JMERE) task to handle the multimodal NER and RE tasks. To tackle this task, we propose an edge-enhanced graph alignment network and a word-pair relation tagging. Specifically, we design a word-pair relation tagging to avoid the error propagation caused by the pipeline framework. Then, we propose an edge-enhanced graph alignment network (EEGA) to enhance the JMERE task by aligning nodes and edges simultaneously in the cross-graph. The EEGA can leverage the edge information to auxiliary alignment between objects and entities and find the correlations between entity-entity relationships and object-object relationships. The detailed evaluation demonstrates that our proposed model significantly outperforms several state-of-the-art baselines. We will extend our approach to multi-label multimodal tasks in our future work and investigate other methods (e.g., the self-supervised model) to better model JMERE.

\section{Acknowledgement}
This work was supported by National Natural Science Foundation of China (62076100, 61966038), and Fundamental Research Funds for the Central Universities, SCUT (x2rjD2220050), the Science and Technology Planning Project of Guangdong Province (2020B0101100002), the CAAI-Huawei MindSpore Open Fund, the Hong Kong Research Grants Council (project no. PolyU 11204919 and project no. C1031-18G) and an internal research grant from the Hong Kong Polytechnic University (project 1.9B0V).
\bibliography{aaai23}

\appendix
\section{Appendix}
\subsection{Appendix A: Detail about the Multi-channel Layer}

In this section, we introduce the process of constructing three feature matrices: part of speech (Pos), syntactic distance
(Sd), and word co-occurrences matrix (Co).

\begin{figure}[ht]    
  \centering
    \includegraphics[scale=.50]{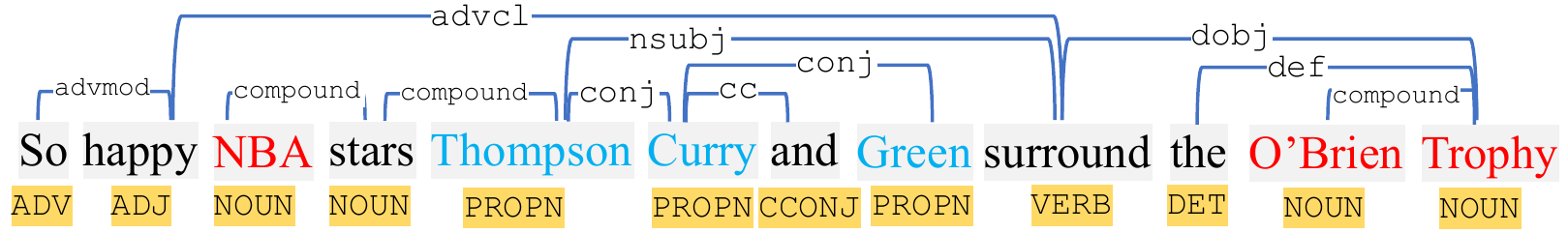}
  \caption{The dependency parse tree of the given example.}
  \label{FIG:dependency}
\end{figure}

\noindent \textbf{Part of Speech (Pos).} To obtain the lexical-level information between word pairs $(w_i,w_j)$, we use the spaCy to obtain the Pos sequence of the input sentence, as shown in Figure~\ref{FIG:dependency}. Then, an embedding layer is used to embed the Pos sequence into trainable vectors $R^{Pos} \in \mathbb{R} {^{n \times {d_l}}}$, . We add the \emph{i}-th to \emph{j}-th Pos vector of the corresponding element as the final lexical-level representation $R_{i,j}^{Pos} \in \mathbb{R} {^{{d_l}}}$.

\noindent \textbf{Syntactic Distance (Sd).} To model the syntactic-level information for word pairs. we use the syntactic relative distance $M_{i,j}^{Sd}\in \mathbb{R} {^{{1}}}$, which is defined as the number of hops on path from the token $x_i$ to token $x_j$ in a dependency tree. As shown in Figure~\ref{FIG:dependency}, the absolute distance between \textbf{Thompson} and \textbf{Trophy} is 7, while the syntactic relative distance is 2. 

\begin{table}[ht]
\centering
\begin{tabular}{|c|c|c|c|c|}

\hline
\multirow{2}{*}{Statistics} & \multicolumn{4}{c|}{JMERE} \\
\cline{2-5}
& \#S & Entity & Words & \#AL \\

\hline
Train                    & 3,618     & 9,006           & 15,981 & 16.31         \\
\hline
Dev                      & 496      & 1,248           & 3,535  & 16.57         \\
\hline
Test                     & 475      & 1,280           & 4,678  & 16.28  \\    
\hline
\end{tabular}

\caption{The statistics of the JMERE dataset. Here \#S and Entity respectively denote the numbers of sentence and entity. The \#AL is the average length of the sentence.
}\label{table:1}
\end{table}
\noindent \textbf{Co-occurrences matrix (Co).} To incorporate the corpus-level information into the word pairs, such as \textbf{Curry} and \textbf{NBA} co-appearing many times in the corpus, we use the word frequency co-occurrence matrix with the Pointwise Mutual Information (PMI) \cite{bouma2009normalized} to calculate the word-pair correlation. Thus, the correlation of PMI for the \emph{i}-th token with the \emph{j}-th token is denoted as $PMI_{i,j}$. To encoding $PMI_{i,j}$ as a trainable vector, we round up the $PMI_{i,j}$ and obtain the final correlation value $M_{i,j}^{Co}\in \mathbb{R} {^{{1}}}$. Especially for negatively correlated values, we uniformly set them as -1.

\subsection{Appendix B: Dataset}

To verify the effectiveness of the proposed model, we collect the joint multimodal entity-relation extraction dataset (JMERE) by the MNER \cite{Zheng2021MEGA} and MRE datasets \cite{Zheng2021}. In addition, we merge samples of the same sentence but with different annotations for entities and relations. Each sample contains the original sentence, corresponding image, and sets of quintuples. If the corpus contains many unlabeled data, the model will learn some meaningless information. Thus, we eliminate the original dataset samples with no entity type or relationship between entities. Table \ref{table:1} shows the static of our collated JMERE.

To evaluate the performance of different methods, we use precision, recall, and F1-score as the evaluation metrics. The extract quintuple is regarded as correct only if predicted, and ground truth spans match precisely.

\subsection{Appendix C: Implementation Details}
We apply the spaCy\footnote{https:github.com/explosion/spaCy} with \emph{en\_core\_web\_trf} version to parse the given sentence into a dependency tree, then built both text graph and adjacency matrix from the dependency tree. We initialize the textual representation by \texttt{BERT-based-Uncased} \footnote{https:huggingface.co/bert-base-uncased} and set the dimension $D_T$ is 768. Besides, the dimension $D_I$ of visual object extracting by scene graph generation model \cite{Tang2020} (the Mask-RCNN used as the backbone) is 4096. In addition, the dimension of the other features, e.g., edge features $d_{zT}$ and $d_{zI}$ and linguistic features $d_l$, are initialized 100. The iteration number of the Sinkhorn algorithm  is 20, and the maximum number of token sequences and objects is 70 and 10, respectively. Adam \cite{Kingma2015} optimizer with a learning rate of 2e-5 and a decay factor of 0.5. The early stopping strategy is also applied to determine the number of epochs with the patience of 5. We implement our model with the PyTorch framework and conduct experiments on the machine with NVIDIA RTX 3090.
\begin{figure}    
  \centering
    \includegraphics[scale=.54]{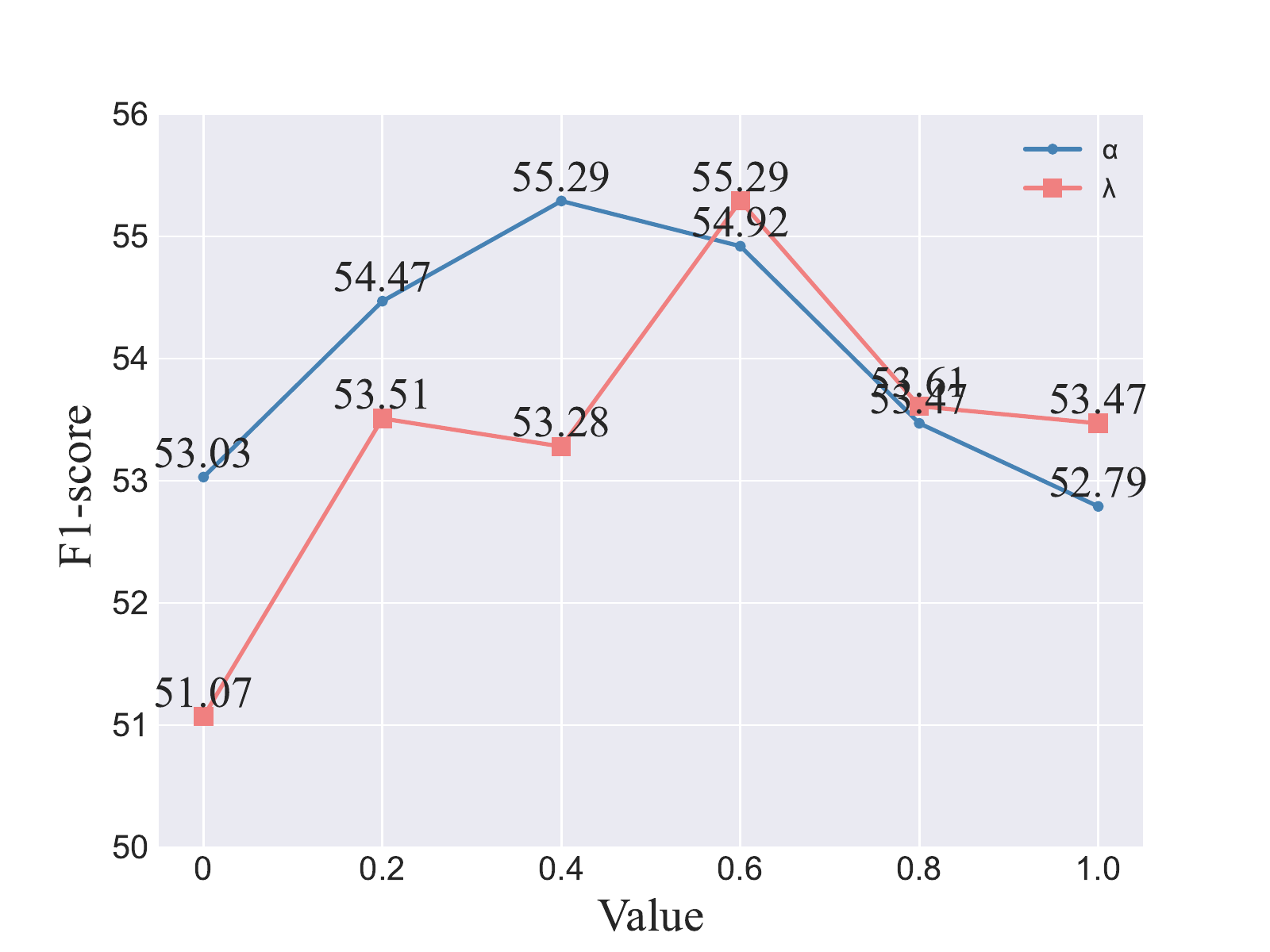}
  \caption{The effect of different parameters about balance coefficients, where $\alpha$ in Eq. (7) and $\lambda$ in Eq. (13).}
  \label{FIG:5}
\end{figure}
\subsection{Appendix D: Parameter Sensitivity}
In this section, we further discuss the different settings of the parameters. We are concerned about the influence of two balance coefficients $\alpha$ in Eq.(7) and $\lambda$ in Eq. (13). The $\lambda=0$ means \emph{w/o edge-enhanced}, the model cannot effectively match the objects with entities and achieves the worst performance. When $\lambda=0.6$ the proposed EEGA achieved the best performance; After $\lambda$ is over 0.6, the cross-graph alignment has interfered with the training process of the main task, resulting in slightly lower performance. In addition, the $\alpha=0$ and $\alpha=1$ mean only aligned nodes and edges in the cross-graph alignment and achieve a lower performance, especially when $\alpha=1$. This shows that edge alignment can effectively improve the precision of matching entities with objects and the ability to capture latent semantic relationships between objects. When $\alpha=0.4$, the result of EEGA is better than other settings.

\end{document}